\documentclass[10pt,twocolumn,letterpaper]{article}
\usepackage{iccv}
\usepackage{times}
\usepackage{epsfig}
\usepackage{graphicx}
\usepackage{amsmath}
\usepackage{amssymb}
\usepackage{multirow}
\usepackage{booktabs}
\usepackage{makecell}
\usepackage{xcolor}

\usepackage{color, colortbl}
\usepackage{makecell}
\usepackage[accsupp]{axessibility}

\usepackage{authblk}

\usepackage[pagebackref=true,breaklinks=true,letterpaper=true,colorlinks,bookmarks=false]{hyperref}

\iccvfinalcopy 

\makeatletter
\def\thanks#1{\protected@xdef\@thanks{\@thanks
        \protect\footnotetext{#1}}}
\makeatother
\begin{document}
\title{Video Action Recognition with Attentive Semantic Units}

\author{
Yifei Chen$^{1}$
~~~~~~Dapeng Chen$^{1\dagger}$\thanks{$\dagger$ Corresponding author. $\star$ Ruijin Liu and Hao Li were interns at Huawei during the project.}
~~~~~~Ruijin Liu$^{2\star}$
~~~~~~Hao Li$^{3\star}$
~~~~~~Wei Peng$^{1}$\\
$^1$ IT Innovation and Research Center, Huawei Technologies Ltd\\
$^2$ Xi’an Jiaotong University~~~~~~~~$^3$ Xiamen University\\
\{chenyifei14, chendapeng8\}@huawei.com
}

\maketitle
\ificcvfinal\thispagestyle{empty}\fi

\begin{abstract}

Visual-Language Models (VLMs) have significantly advanced video action recognition.
Supervised by the semantics of action labels, recent works adapt the visual branch of VLMs to learn video representations.
Despite the effectiveness proved by these works, we believe that the potential of VLMs has yet to be fully harnessed. 
In light of this, we exploit the semantic units (SU) hiding behind the action labels and leverage their correlations with fine-grained items in frames for more accurate action recognition.
SUs are entities extracted from the language descriptions of the entire action set, including body parts, objects, scenes, and motions.
To further enhance the alignments between visual contents and the SUs, we introduce a multi-region attention module (MRA) to the visual branch of the VLM. The MRA allows the perception of region-aware visual features beyond the original global feature. 
Our method adaptively attends to and selects relevant SUs with visual features of frames. With a cross-modal decoder, the selected SUs serve to decode spatiotemporal video representations.
In summary, the SUs as the medium can boost discriminative ability and transferability.
Specifically, in fully-supervised learning, our method achieved 87.8\% top-1 accuracy on Kinetics-400. In K=2 few-shot experiments, our method surpassed the previous state-of-the-art by +7.1\% and +15.0\% on HMDB-51 and UCF-101, respectively.

\end{abstract}

\section{Introduction}
\label{sec:intro}
Video action recognition is the fundamental task toward intelligent video understanding. 
Facilitated by deep learning, great advancements have been made in designing end-to-end network architecture, 
including two-stream networks~\cite{simonyanTwostreamConvolutionalNetworks2014a, wangTSN, zhouTemporalRelationalReasoning2018}, 
3D convolutional neural networks (3D-CNN)~\cite{carreiraQuoVadisAction2017a,feichtenhoferX3dExpandingArchitectures2020, feichtenhoferSlowfastNetworksVideo2019, haraLearningSpatiotemporalFeatures2017, qiuLearningSpatiotemporalRepresentation2017, tranLearningSpatiotemporalFeatures2015, tranCloserLookSpatiotemporal2018, xieRethinkingSpatiotemporalFeature2018} and transformer-based networks~\cite{bertasiusSpacetimeAttentionAll2021, fanMultiscaleVisionTransformers2021, liuVideoSwinTransformer2022a,patrickKeepingYourEye2021, yanMultiviewTransformersVideo2022}. 
The visual-languages models (VLM)\cite{radfordLearningTransferableVisual2021,chaoALIGN21} have achieved great success in the last few years. Taking inspiration from this, recent works\cite{ wangActionclipNewParadigm2021, juPromptingVisualLanguageModels2022,niXCLIP} started exploiting video representation learning under semantic supervision. In particular, these works propose a paradigm that trains video representations to align with action-name text embeddings, which provide richer semantic and contextual information than conventional one-hot labels. Besides showing competitive performance in closed-set action recognition scenarios, adopting VLM demonstrates great learning effectiveness toward recognizing unseen or unfamiliar categories compared to methods that do not utilize text embeddings. 


\begin{figure*}[t]
\centering
\includegraphics[width=0.9\linewidth]
{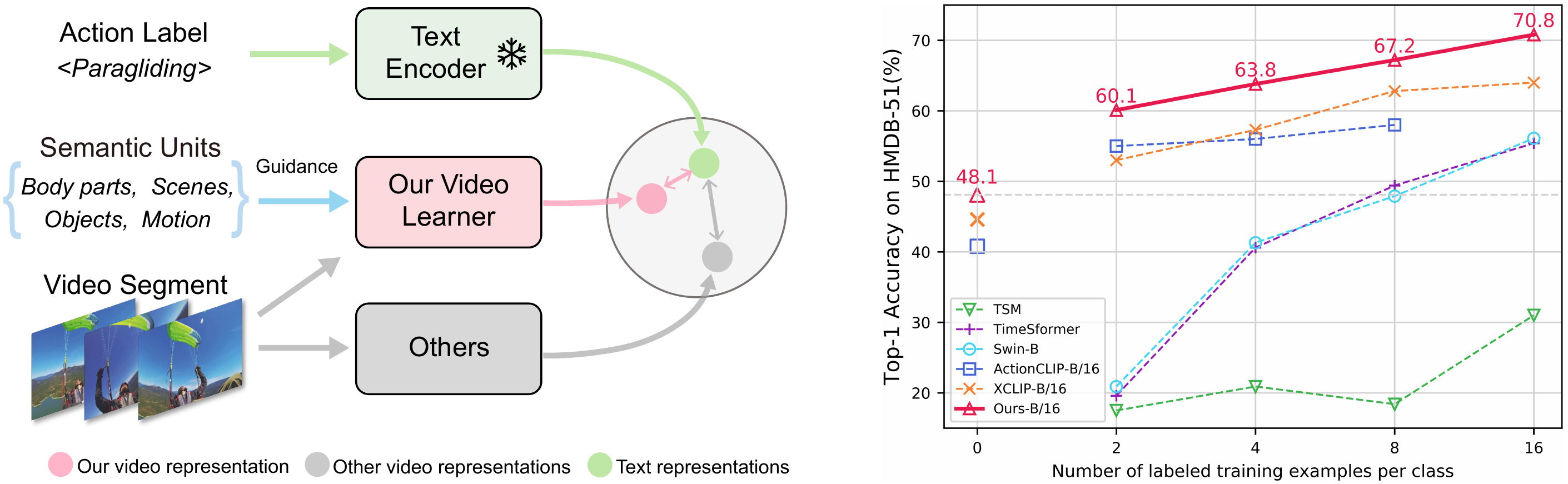}
\caption{
\textbf{Left:} Compared with conventional video learners that encode frames directly, our method utilizes attentive semantic units (SU) to guide the representation learning of videos. The learned representation contains rich semantics and thus can be effectively adapted to the target representation given by the text representation of the action name.
\textbf{Right:}  With the utilization of semantic units (SU), our method can achieve impressive results on zero/few-shot learning. With the model trained on Kinetics-400, we achieve 48.1$\%$ and 70.8$\%$ top-1 accuracy in zero-shot and K=16 few-shot experiments, which is  3.5$\%$ and 6.6$\%$ higher than the second best methods.\vspace{-1em}
}
\label{fig:pipeline_compare}
\end{figure*}

Despite the achieved improvements, we believe that relying solely on the semantic information of action labels is far from enough to fully harness the advantages of the VLMs. This is because action labels are abstract concepts, and directly mapping a video clip to them may confront a noticeable gap. We propose alleviating the problem by relating the action labels with some common entity, ranging from static scenes to moving body parts. For example, ``playing golf" can be associated with people, golf equipment, and grass, which could provide fine-grained and specific guidance for visual alignments. 
We refer to these action-related basic entities as \emph{semantic units} (SU). In practice, SUs are extracted from discriminative language descriptions of the whole action set, including body parts, objects, scenes, and motions.
Our basic idea is to use visual features to select related SUs and then use SUs to guide the decoding of spatiotemporal video features. Compared with the previous approaches, utilizing SUs can bring two advantages:
(1) The text corresponding to semantic units can better explain which factors determine an action.
(2) The re-usability of semantic units can greatly alleviate learning difficulty in the zero-shot/few-shot scenario.

Another concern arises when we explore fine-grained visual-language correlations through SUs: the visual branch of current VLMs produces frame-level representations, which may hinder the sensitive correspondences to SUs. To leverage the local visual appearance, we further introduce multi-region attention (MRA) modules upon the pre-trained visual encoders. MRA spatially perceives sub-regions of each image and enhances the visual features with region-aware info, therefore enabling visual features to select SUs globally and locally.
The final video representation is produced by a cross-modal video decoder with the selected SUs as the queries and the enhanced visual features from frames as the keys and values. The cross-modal decoder is a stack of cross-attention and 1D temporal convolution hybrid modules. 
Cross-attention makes the representation focus on the most critical visual appearance by attentive semantic units, while the temporal convolution exploits the motion cue for video representation.
In summary, our contributions are threefold: 
(1) We utilize semantic units to guide the representation learning of actions. Because the semantic units are fine-grained and reusable, the learned representation can improve both the discriminative ability and the transferability. 
(2) We introduce multi-region attention (MRA) to the visual branch of VLM, which originally provided global representations. MRA perceives region-aware features for each image, enabling sensitive correspondence to multiple fine-grained semantic cues behind actions. 
(3) We propose a cross-modal decoder that generates the final video representation. This module utilizes attentive semantic units to highlight critical visual appearances while exploiting motion cues.

\section{Related Work}

\noindent \textbf{Network Architectures in Action Recognition}
With the significant advances in deep learning, the current video recognition approaches aim to learn effective spatial-temporal representation by well-designed network architectures. The two-stream networks apply 2D CNNs in video recognition~\cite{zhouTemporalRelationalReasoning2018, wangTSN, simonyanTwostreamConvolutionalNetworks2014a}, where they combine the color image and optical flow to obtain inter-frame cues. The availability of larger video classification datasets ,such as Kinetics~\cite{kay_kinetics_2017}, facilitates the training of 3D CNNs~\cite{feichtenhoferX3dExpandingArchitectures2020, tranCloserLookSpatiotemporal2018, qiuLearningSpatiotemporalRepresentation2017, tranLearningSpatiotemporalFeatures2015, xieRethinkingSpatiotemporalFeature2018, feichtenhoferSlowfastNetworksVideo2019, feichtenhoferX3dExpandingArchitectures2020}. 3D CNN factorizes the convolution across spatial and temporal dimensions and can capture temporal variation among the local neighborhood. The recent transformer-based approaches advance the state-of-the-art previously set by 3D CNNs, including ViViT~\cite{arnabVivitVideoVision2021a}, Timesformer~\cite{bertasiusSpacetimeAttentionAll2021}, Video Swim Transformer~\cite{liuVideoSwinTransformer2022a}, and Multi-view Transformer~\cite{yanMultiviewTransformersVideo2022}.  Because of the self-attention modules, the transformer-based approaches can
capture the long-term spatial-temporal dependencies, leading to better performance. 
Our method also employs Transformers as backbones: a VIT-like network structure to extract the frame-and-region-level representations and a decoder with semantic-unit attention.

\begin{figure*}[t]
\centering
\includegraphics[width=0.99\linewidth]{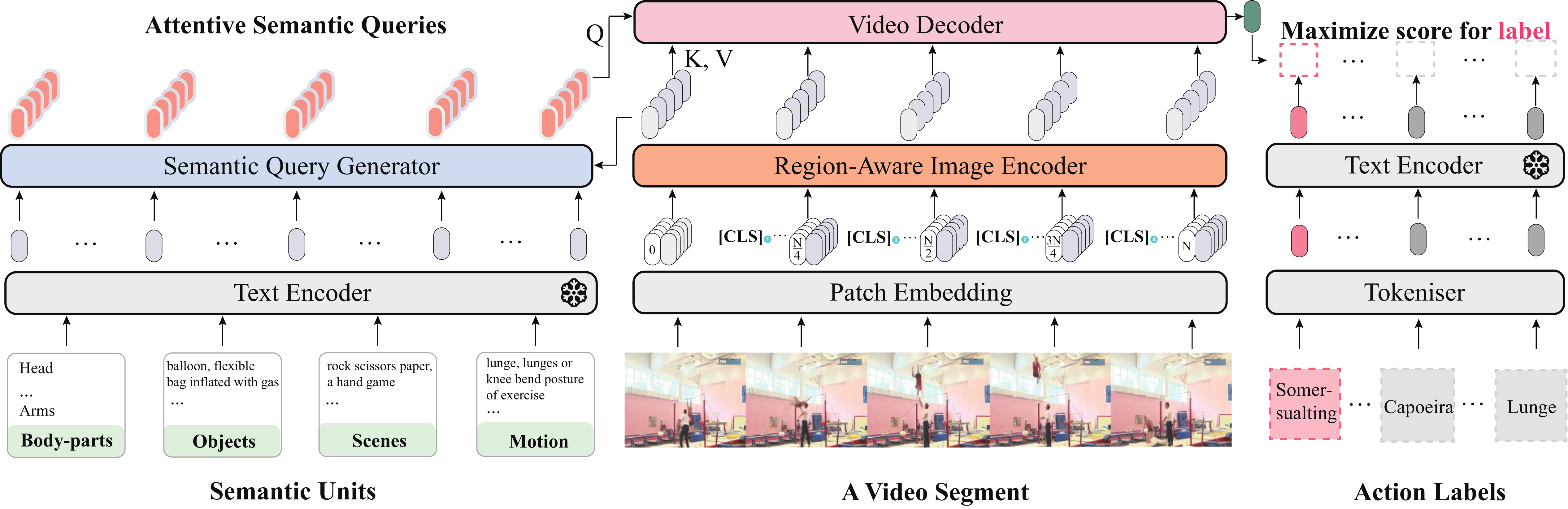}
\caption{An overview of our framework. We input semantic units and a video segment and output a video representation(in dark green). The video representation is supervised by the goal of  
maximizing the similarity score with the text feature of the correct action label.} 
\label{fig:saunet2} \vspace{-1em}
\end{figure*}

\noindent \textbf{Vision-Language Model in Action Recognition} Vision-language models (VLM) are making rapid progress and impressive performance on multiple downstream tasks, including detection~\cite{guOpenvocabularyObjectDetection2021}, segmentation~\cite{wangCrisClipdrivenReferring2022}, caption~\cite{mokadyClipcapClipPrefix2021a}, summarization~\cite{narasimhan_clip-it_2021}, etc. There is also a trend to extend VLMs to video action recognition.  VideoCLIP~\cite{HuVideoCLIP} replaces image-text with video-text pairs to formulate a video-language pretraining for the downstream action recognition task.
However, such an ``training from scratch" strategy is data-hungry and computationally expensive. Another pipeline is to finetune upon the available vision-language pre-trained model such as CLIP~\cite{radfordLearningTransferableVisual2021} and ALIGN~\cite{chaoALIGN21}. 
ActionCLIP~\cite{wangActionclipNewParadigm2021} firstly proposes a ``pre-train, prompt and finetune" framework. 
Prompting-CLIP~\cite{juPromptingVisualLanguageModels2022}  adapts different downstream tasks by prompting learning. 
X-CLIP~\cite{niXCLIP} introduces a cross-frame attention mechanism and utilizes the video content embedding to generate the text prompts, which is used to enhance the text embedding. 
To improve training efficiency,  EVL~\cite{linFrozenCLIPModels2022} and  ST-Adapter~\cite{panSTAdapterParameterEfficientImagetoVideo2022}  fix the CLIP backbone and add a lightweight network module to exploit the spatiotemporal information. 
Unlike all the above methods that leverage semantics by direct mappings, \emph{i.e.} from video embedding to text features of action label names, our approach further utilizes the capability of pre-trained VLMs by introducing fine-grained semantic units to guide the learning of video representations. 

\noindent \textbf{Human-Object Interaction in Action Recognition} The interactions between the object and the human are essential for action recognition.
It was first studied for still images to resolve the ambiguities where two actions have almost identical poses.
Towards action recognition tasks, the HOI methods develop a fine-grained manner based on bounding boxes of objects, humans,  and scenes~\cite{dongCategoryAwareTransformerNetwork2022, fangPairwiseBodyPartAttention2018, gkioxariActionsAttributesWholes2015, liPastanetHumanActivity2020}. For example,  Fang et al.~\cite{fangPairwiseBodyPartAttention2018} explicitly consider the body parts and their pairwise correlations, then fuse the body representations with the scene and objects. The idea of HOI is then applied to the video data by considering the motion cues. 
Something-else~\cite{SomethingElse} mainly works on the dynamic interactions between hands and objects. 
ActionGenome ~\cite{ActionGenome} represents an action by a spatiotemporal scene graph, describing the person-object relationships. 
AKU~\cite{VisualKnowledge} infer actions from body part movements with the assistance of visual-driven semantic knowledge mining. 
Our inspiration for exploring action-related entities comes from the above works. However, in contrast to their reliance on detectors for localization and recognition, our approach efficiently leverages text features of semantic units to select the body parts/instances in frames.

\section{Methodology}
The overall framework of our method is illustrated in Fig.~\ref{fig:saunet2}.
Our method aims to learn more transferable and expressive video representation for action recognition by involving ``semantic units" (SU) for guidance. 
We generate and embed semantic units (Section~\ref{Semantic_Units}), then incorporate the semantic units into the network architecture. 
The overall architecture has three main modules for video representation (Section~\ref{Networks}). 
The training details are introduced in Section~\ref{training_and_testing}. 

\subsection{Semantic Units} \label{Semantic_Units}

Inspired by cognitive science~\cite{kurbySegmentationPerceptionMemory2008,zacksPerceivingRememberingCommunicating2001}, we believe an effective video representation should describe the visual appearance and temporal variation of the body, object, and scenes. 
These factors can be represented by languages and have correspondences in visual appearance. We call them ``semantic units". 

Semantic units are in the form of discriminative descriptions. 
We select nouns or phrases from action names. Then we utilize WordNet~\cite{GeorgeWordNet}, Wikipedia, and a few manual modifications to generate discriminative descriptions to represent the context of the selected texts.
\emph{e.g.}, the word \texttt{polo} is enhanced as \texttt{polo, a game played on horseback with a long-handled mallet} to be distinguished from the meaning \texttt{polo shirt}. Before generation, we remove the stop word in action names via NLTK~\cite{NLTK}. 
On the other hand, we add body-related semantic entities that are sufficiently discriminative by themselves, including \texttt{head}, \texttt{arms}, \texttt{hands}, \texttt{hips}, \texttt{legs}, and \texttt{feet}. They occupy the majority of human activities and are shared by most actions. An illustration of producing semantic units are provided in \textit{supplementary materials}.

\begin{figure*}[t]
\centering
\includegraphics[width=0.99\linewidth]{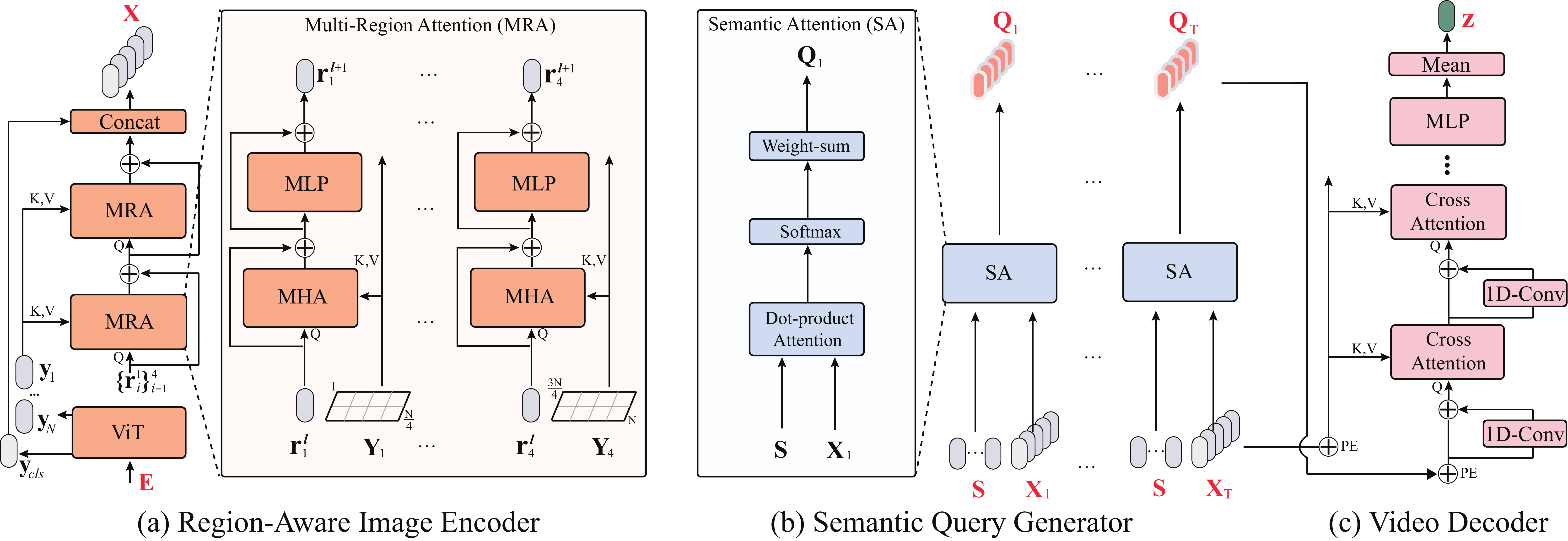}
\caption{Detailed Network Components. Red variables denote the input or output of the region-aware image encoder, semantic query generator, and video decoder.
(a) shows the region-aware image encoder containing a ViT and two multi-region attention (MRA) modules, each of which is shared for four regions. (b) shows the semantic query generator that contains a parameter-free semantic attention (SA) module, and (c) shows the video decoder, which is a stack of hybrid modules composed of cross-attention and 1D temporal convolution.
} 
\label{fig:query_gen_vid_dec} \vspace{-1em}
\end{figure*}

All the semantic units are collected in the set $\mathcal{S}=\{ s_{i}\}_{i=1}^{K}$. 
We manually categorized them into ``body parts",  ``objects",  and ``scenes". 
For a minority of valid words that solely depend on body motions, such as \texttt{sidekick} and \texttt{squat}, we categorize them as motion units. Therefore,
\begin{equation}
    \mathcal{S} = \{  \mathcal{S}^{body}, \mathcal{S}^{object}, \mathcal{S}^{scene}, \mathcal{S}^{motion} \}. \label{action:labels}
\end{equation}
We employ the text encoder of CLIP to encode every semantic unit and its elaborative description to feature vectors. 
The feature vectors of all the semantic units are concatenated to be $\mathbf{S} \in \mathbb{R}^{K\times d}$, where $K$ is the number of semantic units, and $d$ is the dimension of the text feature.

\subsection{Network Architecture} \label{Networks}

Given a video segment with $T$ frames,  we extract the visual representations $\mathbf{X}$ for each frame by a region-aware image encoder, and
utilize a semantic query generator to produce the queries $\mathbf{Q}$.
The visual representations and queries of all the frames, $\widehat{\mathbf{X}}$ and $\widehat{\mathbf{Q}}$, are fed to the video decoder, which outputs the video representation $\mathbf{z}$.


\noindent \textbf{Region-Aware Image Encoder}
The region-aware image encoder, as shown in Fig.~\ref{fig:query_gen_vid_dec}a, takes the video frame as input and provides the frame-level and region-level representations. The encoder consists of two parts: a ViT initialized by CLIP is used to generate frame-level representations; Two consecutive multi-region attention (MRA) modules are used to generate the region-level representations. 

\begin{figure}[t]
\centering
\includegraphics[width=0.95\linewidth]{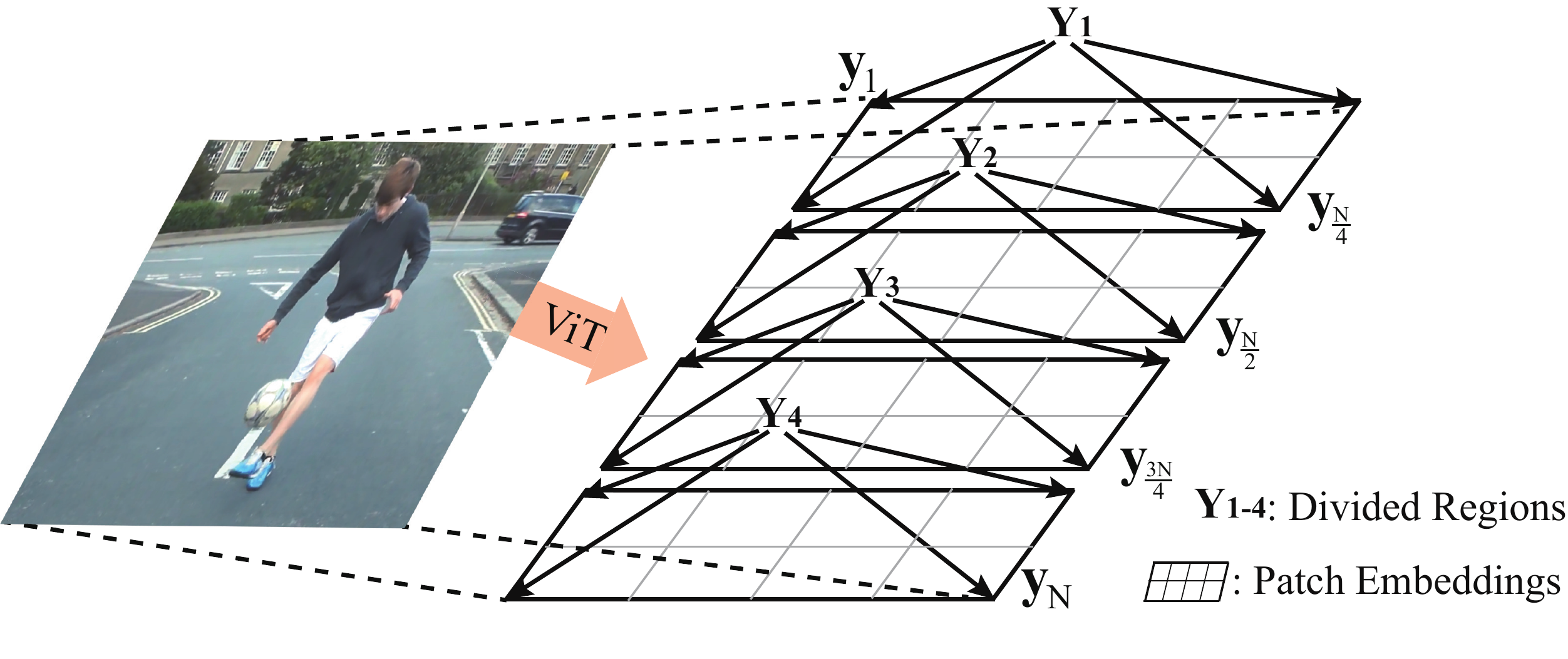}
\caption{Visualization of dividing patch representations into regions. Note that when patches are not divisible by four along the vertical axis, we assign more lines of patches to middle regions.
} 
\label{fig:region_division} \vspace{-1em}
\end{figure}

For a frame $\mathbf{I} \in \mathbb{R}^{H \times W \times 3}$, we divide the image into  $N$ non-overlapping
patches following ViT. The patches are then embedded into vectors $\{\mathbf{e}_{i}\}_{i=1}^{N}$.
A learnable embedding $\mathbf{e}_{cls}$ is prepended to the sequence of the embedded patches, called $[class]$ token. The embeddings are concatenated to form a matrix, which is further added with the position embeddings $\mathbf{E}_{pos}$ to obtain $\mathbf{E}\in \mathbb{R}^{(N+1)\times d}$:
\begin{equation}
 \textbf{E} = [\textbf{e}_{cls}^{\top}, \textbf{e}_{1}^{\top}, \textbf{e}_{2}^{\top}, ..., \textbf{e}_{N}^{\top}]^{\top} + \textbf{E}_{pos}.
 \end{equation}
 We feed $\textbf{E}$ to ViT, producing the visual representations $\mathbf{Y}$:
\begin{equation}
 \textbf{Y} = [\textbf{y}_{cls}^{\top}, \textbf{y}_{1}^{\top}, \textbf{y}_{2}^{\top}, ..., \textbf{y}_{N}^{\top}]^{\top},
\end{equation} 
where $\textbf{y}_{cls} \in \mathbb{R}^{d}$ is the frame-level representation according to ViT, and $(\textbf{y}_{1}, \textbf{y}_{2}, ..., \textbf{y}_{N})$, obtained by averaging the patch embeddings of the last two layers of the ViT, are the representations for the corresponding patches.
Inspired by strategy from the re-ID community ~\cite{Chen_reID_2016, EomReID2019}, the N patch representations are split into groups based on their affiliation to the four vertically arranged equal-size regions, forming $\{\mathbf{Y}_{i}\}_{i=1}^{4}$. The procedure is illustrated in Fig.~\ref{fig:region_division}.  
Utilized as the keys and values, patch representations are further fed into an L-layer cross-attention (MRA) module, where each group $\mathbf{Y}_{i}$ of them serves the generation of the corresponding region-level representation:
\begin{equation}
\mathbf{r}_{i}^{l} = {\rm MRA}^{l}(\mathbf{r}_{i}^{l-1}, \mathbf{Y}_{i}), \quad l=1,...,L. 
\end{equation}
In particular, $l$ denotes the index of MRA and $L$ = 2, $\mathbf{r}_{i}^{1}$ are learnable embedding for region $i$ with positional embeddings added, and $\mathbf{r}_{i}^{L}$ is region-level representation we finally obtain. MRA consists of Multi-Head Attention (MHA) and Multi-Layer Perceptron (MLP), thus can be represented by:
\begin{equation}
\begin{split}
    &{\rm MRA}(a,b) = c + {\rm  MLP}(c),\\
    &c = a + {\rm MHA}(a,b,b). 
\end{split}   
\label{eq:CA}
\end{equation}
It is noteworthy that $\{\mathbf{r}_{i}^{l}\}_{i=1}^{4}$ are parallelly computed by ${\rm MRA}^{l}$ via an attention mask. The final output of the region-aware image encoder is:
\begin{equation}
    \mathbf{X} = [ (\mathbf{y}_{cls})^{\top}, (\mathbf{r}_{1}^{L})^{\top},
    (\mathbf{r}_{2}^{L})^{\top},
    (\mathbf{r}_{3}^{L})^{\top},
    (\mathbf{r}_{4}^{L})^{\top} 
    ]^{\top}. \label{eq:featmap}
\end{equation}
$\mathbf{X}\in \mathbb{R}^{5 \times d}$ combines the frame-level representation and the region-level representations.


\vspace{0.2em}

\noindent \textbf{Semantic Query Generator}
We generate query feature $\mathbf{Q}$ for a frame with $\mathbf{S}$ and $\mathbf{X}$ as input.
As illustrated in Fig.~\ref{fig:query_gen_vid_dec}b, we feed the representations  $\mathbf{X}$ to Semantic Attention (SA), a parameter-free block allows each visual representation in $\mathbf{X}$ to select the semantic units. 
\begin{equation} \label{eq:cosine_sim}
w_{i,j}=\frac{\langle \mathbf{x}_i, \mathbf{s}_j\rangle} {\parallel \mathbf{x}_i \parallel \cdot \parallel \mathbf{s}_{j} \parallel}, \quad  
 a_{i,j}= \frac
{\exp(w_{i,j}/\tau)}
{\sum_{j=1}^K \exp(w_{i,j}/\tau)},
\end{equation}
where $\mathbf{x}_{i} \in \mathbb{R}^{d}$ and $\mathbf{s}_{j} \in \mathbb{R}^{d}$ are $i$th row and $j$th row of $\mathbf{X}$ and $\mathbf{S}$. $w_{i,j}$ is cosine similarity between $\mathbf{x}_{i} $ and $\mathbf{s}_{j}$. 
With the attention weights $a_{i,j}$ calculated by the softmax activation of $w_{i,j}$   over the $K$-class semantic units $\mathbf{S}$ under temperature $\tau$,
we obtain the affinities matrix $\mathbf{A} \in \mathbb{R}^{5\times K}$ between the representations  $\mathbf{X}$ and $\mathbf{S}$.
The query features $\mathbf{Q} \in \mathbb{R}^{5\times d}$ for frame $\mathbf{I}$ is then calculated in a weight-sum manner:
\begin{equation}
    \mathbf{Q} =  \mathbf{A}\mathbf{S}. \label{eq:query}
\end{equation}

\vspace{0.2em}

\noindent \textbf{Video Decoder} Given a video $[\mathbf{I}_{1}, \mathbf{I}_{2},...,\mathbf{I}_{T}]$, we generate the feature maps and queries for each frame according to Eq.~\ref{eq:featmap} and Eq.~\ref{eq:query}, obtaining $\widehat{\mathbf{X}}  \in \mathbb{R}^{T \times 5 \times d}$ and $\widehat{\mathbf{Q}}\in \mathbb{R}^{T \times 5 \times d}$:
\begin{equation}
    \begin{split}
        \widehat{\mathbf{X}} \!=\! [\mathbf{X}_{1},..., \mathbf{X}_{i},...,\mathbf{X}_{T}], \quad
          \widehat{\mathbf{Q}} \!=\! [\mathbf{Q}_{1},..., \mathbf{Q}_{i},...,\mathbf{Q}_{T} ].
    \end{split}
\end{equation}
The video decoder, as shown in Fig.~\ref{fig:query_gen_vid_dec}c, 
consists of a sequence of a 1D-convolution layer and a cross-attention (CA) module, whose processing mechanism is similar to Eq.~\ref{eq:CA}.
The 1D-convolution layer processes the queries along the temporal dimension, while the CA module utilizes the representation of the queries to attend to the visual feature across the frames. 
It is noteworthy that the input queries to the video decoder 
are derived from the weighted sum of SUs' text embeddings, while the keys and values are from the region-aware image encoder. The video decoder promotes communication between the two modalities. More specifically, we perform the cross attention for each region and denote the visual representations and queries for the $i$th region as $\widehat{\mathbf{X}}_{i} \in \mathbb{R}^{T \times d}$ and $\widehat{\mathbf{Q}}_{i} \in \mathbb{R}^{T \times d }$. The procedure of the video decoder is as follows:
\begin{equation}
    \begin{split}
     \widehat{\mathbf{H}}_{i}^{m} & =  \widehat{\mathbf{Q}}_{i}^{m-1} + {\rm 1Dconv}(\widehat{\mathbf{Q}}_{i}^{m-1} ) \\ 
     \widehat{\mathbf{Q}}_{i}^{m}  & =   {\rm CA}^{m}(\widehat{\mathbf{H}}_{i}^{m}, \widehat{\mathbf{X}}_{i}), \quad m = 1,...,M.
    \end{split}
\end{equation}
where $m$ indicates the layer index of the decoder and $M$ = 4. 
The initialized query $\widehat{\mathbf{Q}}_{i}^{0} = \widehat{\mathbf{Q}}_{i}$, and the output query after $M$-times attention is $\widehat{\mathbf{Q}}_{i}^{M}$. 
We concatenate the queries of different regions along the feature channel, obtaining $\widehat{\mathbf{Q}}^{M} \in \mathbb{R}^{T\times 5d}$. 
The video representation $\mathbf{z}$ is a vector calculated by applying MLP and MEAN over $\widehat{\mathbf{Q}}^{M}$ sequentially along the feature channel and the temporal dimension, 
\begin{equation}
    \mathbf{z} = {\rm  MEAN}({\rm MLP}(\widehat{\mathbf{Q}}^{M})).
\end{equation}

\subsection{Training Details} \label{training_and_testing}
\noindent \textbf{Loss function}  We supervise the video representation with the text features of the action labels. Specifically, we feed action names with a prefixed template ``a video of a person" into the frozen text encoder initialized by CLIP. The resulting text feature of the $i$th action is $\mathbf{c}_{i}$. 
The goal of the proposed network is to maximize the similarity between $\mathbf{z}_{n}$ and $\mathbf{c}_{i}$ if $\mathbf{z}$ belongs to the $i$th action. The loss function can be implemented by the cross-entropy loss:
\begin{equation}
    \mathcal{L} = -\frac{1}{N}\sum_{n=1}^{N}\sum_{i=1}^{I} y^{i,n} \log \left( \frac{\exp(\mathbf{c}_{i}^{\top}\mathbf{z}_{n})}{\sum_{j=1}^{I} \exp(\mathbf{c}_{j}^{\top}\mathbf{z}_{n})} \right),
\end{equation}
where $\mathbf{z}_{n}$ the representation of the $n$th video. The training set has $N$ videos belonging to the $I$ actions. If the $n$th video belongs to the $i$th action, $y^{i,n}=1$, otherwise $y^{i,n}=0$.

\noindent \textbf{Network Training}
The ViT backbone in the region-aware image encoder is initialized by CLIP, while the other parameters in the region-aware image encoder and video decoder are randomly initialized. 
For network parameter training, we adopt an AdamW optimizer with an initial learning rate of 8$\times10^{-6}$ and 8$\times10^{-5}$ for the ViT backbone and the remaining parts, respectively. 
We train the network with 30 epochs and a weight decay of 0.001. 
The learning rate is warmed up for 5 epochs and decayed w.r.t. a cosine schedule. 
The input video follows the main sparse sampling method~\cite{wangTSN} and augmentation strategy~\cite{niXCLIP} with a frame resolution 224$\times$224. 
The hyperparameter temperature of softmax in the semantic query generator is set to 0.01. All the experiments are conducted with 8 32GB-V100-GPUs. More details are provided in \textit{supplementary materials}.



\section{Experiments}
\label{sec:experiments}
\paragraph{Datasets} We benchmark our method on four popular video action recognition datasets. 
(1) Kinetics-400~\cite{kay_kinetics_2017} is a collection of around 240k training and 20k validation videos for 400 classes. Each clip covers around 10s.
(2) Kinetics-600~\cite{joaok600} is an extension from Kinetics-400, which consists of around 410k training and 29k validation videos for 600 classes.
(3) UCF-101~\cite{KhurramUCF101} contains 13,320 video clips with 101 classes. There are three splits of test data. 
(4) HMDB-51~\cite{HildegardHMDB51} consists of 7,000 videos with 51 classes and has three test data splits.
We conduct fully-supervised experiments on Kinetics-400 and Kinetics-600. With the pre-trained model on Kinetics-400, we conduct few-shot and zero-shot experiments on UCF-101 and HMDB-51.




\definecolor{brilliantlavender}{rgb}{0.96, 0.73, 1.0}
\definecolor{celadon}{rgb}{0.67, 0.88, 0.69}
\definecolor{columbiablue}{rgb}{0.61, 0.87, 1.0}

\definecolor{brilliantlavender}{rgb}{0.96, 0.73, 1.0}
\definecolor{celadon}{rgb}{0.67, 0.88, 0.69}
\definecolor{columbiablue}{rgb}{0.61, 0.87, 1.0}
\definecolor{azure(web)(azuremist)}{rgb}{0.94, 1.0, 1.0}
\definecolor{lavender(web)}{rgb}{0.9, 0.9, 0.98}
\definecolor{lavenderblue}{rgb}{0.8, 0.8, 1.0}
\definecolor{lavendermist}{rgb}{0.92, 0.9, 0.98}
\definecolor{mistyrose}{rgb}{1.0, 0.89, 0.88}
\definecolor{bubbles}{rgb}{0.89, 0.98, 1.0}

\begin{table*}
\caption{Comparison with state-of-the-art on Kinetics-400. FLOPs and throughput are reported per view with tools provided by X-CLIP~\cite{niXCLIP}. For multi-modal approaches, parameters in the text branch are not counted. \** indicates pretraining with a video-text collection.}
\label{tab:k400}
\centering
\begin{tabular}{lcccccccc}
\toprule
Method & 
Pretrain & Frames & Top-1 & Top-5 & Views &  \#Param.(M)   & FLOPs(G) & Throughput \\
\midrule
\multicolumn{9}{l}{\textit{Methods with ImageNet or web-scale image pretraining}} \\
MViTv1-B, 64$\times$3~\cite{fanMultiscaleVisionTransformers2021} 
& -      & 64     & 81.2  & 95.1  & 3$\times$3 & 36.6 & 455   & 7   \\
Uniformer-B~\cite{kunchangUniFormer}
& IN-1k  & 32     & 83.0  & 95.4  & 4$\times$3 & 50.0 & 259   & -   \\
TimeSformer-L~\cite{bertasiusSpacetimeAttentionAll2021}
& IN-21k & 96     & 80.7  & 94.7  & 1$\times$3 & 121.4 & 2380  & 3   \\
Mformer-HR~\cite{patrickKeepingYourEye2021}
& IN-21k & 16     & 81.1  & 95.2  & 10$\times$3 & 381.9 & 959  & -   \\
Swin-L~\cite{liuVideoSwinTransformer2022a}
& IN-21k & 32     & 83.1  & 95.9  & 4$\times$3 & 200.0 & 604   & 6   \\
Swin-L(384$\uparrow$)~\cite{liuVideoSwinTransformer2022a}
& IN-21k & 32     & 84.9  & 96.7  & 10$\times$5 & 200.0 & 2107 & -   \\
MViTv2-L(312$\uparrow$)~\cite{YanghaoMviTv2}
& IN-21k & 64     & 81.2  & 95.1  & 3$\times$3 & 218.0 & 455   & 7   \\
ViViT-H/16~\cite{arnabVivitVideoVision2021a}
& JFT-300M & 32   & 84.9  & 95.8  & 4$\times$3 & 647.5 & 8316 & -   \\
MTV-H/16~\cite{yanMultiviewTransformersVideo2022}
& JFT-300M  & 32   & 85.8  & 96.6  & 4$\times$3  & - & 3705 & -   \\
\midrule
\multicolumn{9}{l}{\textit{Methods with web-scale language-image pretraining}}  \\
\rowcolor{lavendermist}
MTV-H/16~\cite{yanMultiviewTransformersVideo2022}
& WTS-17B*   & 32   & 89.1  & 98.2  & 4$\times$3  & - & 3705 & -   \\
PromptingCLIP-B/16~\cite{juPromptingVisualLanguageModels2022}
& \multirow{8}{*}{\rotatebox{270}{CLIP-400M}} & 16   & 76.9  & 93.5  & 5$\times$5 & 95.5 & -    & -   \\
ActionCLIP-B/16~\cite{wangActionclipNewParadigm2021}
& ~ & 16   & 82.6  & 96.2  & 10$\times$3 & 105.2 & 282  & -   \\
ActionCLIP-B/16~\cite{wangActionclipNewParadigm2021}
& ~ & 32   & 83.8  & 97.1  & 10$\times$3 & 105.2 & 563  & -   \\
EVL-B/16~\cite{linFrozenCLIPModels2022}
& ~ & 32   & 84.2  & 96.6  & 3$\times$1  & 114.9 & 1777 & - \\
EVL-L/14~\cite{linFrozenCLIPModels2022}
& ~ & 32   & 87.3  & 97.6  & 3$\times$1  & 357.9 & 2696 & - \\
EVL-L/14(336$\uparrow$)~\cite{linFrozenCLIPModels2022}
& ~ & 32   & 87.7  & 97.8  & 3$\times$1  & 357.9 & 6065 & - \\

X-CLIP-B/16~\cite{niXCLIP}
& ~ & 8    & 83.8  & 96.7  & 4$\times$3  & 131.7 & 145  & 33 \\
X-CLIP-B/16~\cite{niXCLIP}
& ~ & 16   & 84.7  & 96.8  & 4$\times$3  & 131.7 & 287  & 17 \\
X-CLIP-L/14~\cite{niXCLIP}
& ~ & 8    & 87.1  & 97.6  & 4$\times$3  & 451.2 & 658  & 8 \\
X-CLIP-L/14(336$\uparrow$)~\cite{niXCLIP}
& ~ & 16   & 87.7  & 97.4  & 4$\times$3  & 451.2 & 3086 & 2 \\
\midrule
ASU-B/16~(ours)
& CLIP-400M & 8    & 84.1  & 96.3  & 4$\times$3  & 140.2 & 146  & 30 \\
ASU-B/16~(ours)
& CLIP-400M & 16   & 84.8  & 96.7  & 4$\times$3  & 140.2 & 288  & 14 \\
ASU-L/14~(ours)
& CLIP-400M & 8    & 87.8  & 97.8  & 4$\times$3  & 425.3 & 660  & 8 \\
\rowcolor{bubbles}
ASU-L/14(336$\uparrow$)~(ours)
& CLIP-400M & 16   & 88.3  & 98.0  & 4$\times$3  & 425.3 & 3084 & 2 \\
\bottomrule
\end{tabular}
\end{table*}

\begin{table}
\caption{Comparison with state-of-the-art on Kinetics-600.}
\label{tab:k600}
\footnotesize
\centering
\begin{tabular}{p{0.33\linewidth}p{0.16\linewidth}<{\centering}p{0.07\linewidth}<{\centering}p{0.08\linewidth}<{\centering}p{0.07\linewidth}<{\centering}}
\toprule
Method & 
Pretrain & Frames & Top-1  & Views      \\
\midrule
MViT-B-24, 32$\times$3~\cite{fanMultiscaleVisionTransformers2021} 
& -      & 32     & 83.8    & 5$\times$1  \\
Swin-L(384$\uparrow$)~\cite{liuVideoSwinTransformer2022a}
& IN-21k & 32     & 86.1    & 10$\times$5 \\
ViViT-H/16x2 320~\cite{arnabVivitVideoVision2021a}
& JFT-300M & 32   & 83.0    & 4$\times$3 \\
ViViT-H/16x2~\cite{arnabVivitVideoVision2021a}
& JFT-300M & 32   & 85.8    & 4$\times$3 \\
TokenLearner-L/10~\cite{MichaelTokenLearner}
& JFT-300M & 32   & 86.3    & 4$\times$3 \\
Florence(384$\uparrow$)~\cite{LuFlorence}
& FLD-900M & 32   & 87.8    & 4$\times$3 \\
CoVeR~\cite{BowenCoVeR}
& JFT-3B   & 96   & 87.9    & 1$\times$3 \\
\midrule
\rowcolor{lavendermist}
MTV-H~\cite{yanMultiviewTransformersVideo2022}
& WTS-17B  & 32   & 89.6    & 4$\times$3 \\
X-CLIP-B/16~\cite{niXCLIP}
& \multirow{5}{*}{\rotatebox{270}{CLIP-400M}} & 8    & 85.3   & 4$\times$3  \\
X-CLIP-B/16~\cite{niXCLIP}
& ~ & 16   & 85.8     & 4$\times$3  \\
X-CLIP-L/14~\cite{niXCLIP}
& ~ & 8    & 88.3     & 4$\times$3  \\
ASU-B/16~(ours)
& ~ & 8    &  85.7    & 4$\times$3  \\
ASU-B/16~(ours)
& ~ & 16   & 86.4   & 4$\times$3  \\
\rowcolor{bubbles}
ASU-L/14~(ours)
& ~ & 8    & 88.6    & 4$\times$3  \\
\bottomrule
\end{tabular}
\vspace{-0.8em}
\end{table}

\subsection{Fully Supervised Comparison}

\noindent \textbf{Settings}  We conduct the fully-supervised experiments on Kinetics-400 and Kinetics-600. 
Each video is sampled with 8 or 16 frames. 
The multi-view inference is with 3 spatial crops and 4 temporal clips. Three variants of the proposed network, \emph{i.e.}, ASU-B/16, ASU-L/14, and ASU-L/14-336, adopt ViT-B/16, ViT-L/14, and ViT-L/14-336 for the region-aware image encoder, respectively.

\noindent \textbf{Results} 
In Tab.~\ref{tab:k400}, we compare with the state-the-of-art methods on Kinetics-400. 
Under the standard 224$\times$224 input resolution setting, our method achieves the second place \emph{87.8\%}, which is \emph{0.5\%} higher top-1 accuracy than the third-ranked method EVL~\cite{linFrozenCLIPModels2022}. 
It is noteworthy that MTV-H~\cite{yanMultiviewTransformersVideo2022} performs the best by adopting much more pre-training data ( 70M video-text pairs with about 17B images~\cite{niXCLIP}) and a larger pre-trained model(Vit-H). 
Our method outperforms MTV-H pre-trained on JFT-300M by \emph{2.0\%} and achieves the best results among the CLIP-based methods.
Under 336$\times$336 input resolution setting, our method achieves the best Top-1 accuracy of \emph{88.3\%}, which improves upon the previous highest result (X-CLIP~\cite{niXCLIP} and EVL~\cite{linFrozenCLIPModels2022}) by \emph{0.6\%}. 
Tab.~\ref{tab:k600} presents the results on Kinetics-600. 
Our ASU outperforms X-CLIP by \emph{0.4}\% and \emph{0.6}\% by using 8 and 16 frames per video, respectively. 
Moreover, our ASU achieves \emph{88.6}\% with 4 $\times$ fewer input frames and 42.5 $\times$ fewer pre-training data compared to the current state-of-the-art method MTV-H. Our method achieves superior results among the methods that adopt similar-level pre-trained models. We attribute the effectiveness to the proposed visual-semantic attention mechanism, where the attentive semantic units can help exploit critical visual information for learning discriminative video representation.

\begin{table*}
\caption{Few-shot comparisson on HMDB-51 and UCF-101.}
\label{tab:fewshot}
\centering
\begin{tabular}{lccccccccccc}
\toprule
\multirow{2}{*}{Method} & \multirow{2}{*}{Frames} & \multicolumn{4}{c}{HMDB-51} & \multicolumn{4}{c}{UCF-101}  \\
& & $K$=2 & $K$=4 & $K$=8 & $K$=16 & $K$=2 & $K$=4 & $K$=8 & $K$=16 \\
\midrule
TSM~\cite{JiTSM}
& 32 & 17.5  & 20.9  & 18.4   & 31.0  & 25.3  & 47.0  & 64.4 & 61.0 \\
TimeSformer~\cite{bertasiusSpacetimeAttentionAll2021} 
& 32 & 19.6  & 40.6  & 49.4   & 55.4  & 48.5  & 75.6  & 83.7 & 89.4 \\
Swin-B~\cite{liuVideoSwinTransformer2022a} 
& 32 & 20.9  & 41.3  & 47.9   & 56.1  & 53.3  & 74.1  & 85.8 & 88.7 \\
\midrule
ActionCLIP~\cite{wangActionclipNewParadigm2021} 
& 8  & 55.0  & 56.0  & 58.0   & -     & 80.0  & 85.0  & 89.0 & - \\
X-CLIP-B/16~\cite{niXCLIP} 
& 32 & 53.0  & 57.3  & 62.8   & 64.0  & 76.4  & 83.4  & 88.3 & 91.4 \\
X-Florence~\cite{niXCLIP}  
& 32 & 51.6  & 57.8  & 64.1   & 64.2  & 84.0  & 88.5  & 92.5 & 94.8 \\
ASU-B/16~(ours)
& 8  & \textbf{57.7} & \textbf{60.9} & \textbf{65.8} & \textbf{70.1} & \textbf{88.8} & \textbf{92.3} & \textbf{94.0} & \textbf{95.3}    \\
ASU-B/16~(ours)
& 32  & \textbf{60.1} & \textbf{63.8} & \textbf{67.2} & \textbf{70.8} & \textbf{91.4} & \textbf{94.6} & \textbf{96.0} & \textbf{97.2}    \\
\bottomrule
\end{tabular}

\end{table*}

\subsection{Few-shot Comparisons}

\noindent \textbf{Settings} 
The few-shot experiments are evaluated on  HMDB-51 and UCF-101. 
The training set is constructed by randomly sampling 2,4,8, and 16 videos from each class and setting the frame number in each video to 8 or 32. 
Following~\cite{niXCLIP}, we adopt the first split of the test set as an evaluation set and report the result of a single-view inference. 

\noindent \textbf{Results} 
In Tab.~\ref{tab:fewshot}, we present the results of $K$-shot learning. 
Compared with methods employing single-modality pre-trained models,  our method outperforms them by a large margin. 
\emph{e.g.},  when K=2, it improves Swin-B by  \emph{39.2\%} on HMDB-51 and \emph{38.1\%} on UCF-101, demonstrating the effectiveness of the semantic knowledge in the few-shot settings. 
Moreover, our method still achieves state-of-the-art performance compared with the methods that employed the cross-modal pre-trained model. 
On HMDB-51, our method with eight frames per video achieves \emph{57.7\%} top-1 accuracy under $K$=2-shot learning, which is higher than the previous best ActionCLIP with \emph{2.7\%}. 
On UCF-101, our method achieves \emph{91.4\%} top-1 accuracy under $K$=2-shot learning, which outperforms X-Florence by \emph{7.4\%}. Our method has achieved consistent superiority  from $K$=2 to $K$=16, indicating the effectiveness of semantic units.


\begin{table}[t]
\caption{Zero-shot comparison on HMDB-51 and UCF-101.} \label{tab:zeroshot}
\centering
\begin{tabular}{lccc}
\toprule
Method 
& HMDB-51      & UCF-101  \\
\midrule
MTE~\cite{XunMTE} 
& 19.7$\pm$1.6 & 15.8$\pm$1.3 \\
ASR~\cite{QianASR} 
& 21.8$\pm$0.9 & 24.4$\pm$1.0 \\
ZSECOC~\cite{JieZSECOC} 
& 22.6$\pm$1.2 & 15.1$\pm$1.7 \\
UR~\cite{YiUR} 
& 24.4$\pm$1.6 & 17.5$\pm$1.6 \\
TS-GCN~\cite{JunyuTSGCN} 
& 23.2$\pm$3.0 & 34.2$\pm$3.1 \\
E2E~\cite{BiagioE2E} 
& 32.7         & 48 \\
ER-ZSAR~\cite{ShizheERZSAR} 
& 35.3$\pm$4.6 & 51.8$\pm$2.9 \\
\midrule
ActionCLIP~\cite{wangActionclipNewParadigm2021} 
& 40.8$\pm$5.4 & 58.3$\pm$3.4 \\
X-CLIP-B/16~\cite{niXCLIP} 
& 44.6$\pm$5.2 & 72.0$\pm$2.3 \\
ASU-B/16~(ours)
& \textbf{48.1}$\pm$\textbf{2.8} & \textbf{75.0}$\pm$\textbf{3.7} \\
\bottomrule
\end{tabular}
\vspace{-1em}
\end{table}

\subsection{Zero-shot Comparisons}

\noindent \textbf{Settings} 
We evaluate the zero-shot performance of ASU-B/16, which is pre-trained on Kinetics-400 with 8-frame videos. The protocol is the same as in \cite{niXCLIP}:  For HMDB-51 and UCF-101, we average the results on the given three splits and report the top-1 accuracy and standard deviation. For Kinetics-600, we randomly selected 160 categories from 220 new categories, exclusive from  Kinetics-600, three times. Then we report the averaged top-1 and top-5 accuracies and their standard deviations.

\noindent \textbf{Results} We report the zero-shot results in Tab.~\ref{tab:zeroshot} and Tab.~\ref{tab:zeroshotk600}. With the model trained on Kinetics-400, our method outperforms X-CLIP-B/16 by  \emph{3.5\%} and \emph{3.0\%} top-1 accuracy on  HMDB-51 and UCF-101, respectively. When tested on a larger scale dataset  Kinetics-600,  our method outperforms X-CLIP-B/16 by \emph{2.4\%} top-1 accuracy. We attribute the success of our method on the few-shot/zero-shot setting to the utilization of semantic units, whose re-usability alleviates the difficulty of adapting our model to a new scenario.

\begin{table}[t]
\caption{Zero-shot comparison on Kinetics-600.} \label{tab:zeroshotk600}
\centering
\begin{tabular}{lccc}
\toprule
Method 
& Top-1        & Top-5        \\
\midrule
DEVISE~\cite{AndreaDEVISE} 
& 23.8$\pm$0.3 & 51.0$\pm$0.6 \\
ALE~\cite{ZeynepALE} 
& 23.4$\pm$0.8 & 50.3$\pm$1.4 \\
SJE~\cite{ZeynepSJE} 
& 22.3$\pm$0.6 & 48.2$\pm$0.4 \\
ESZSL~\cite{BernardinoESZSL} 
& 22.9$\pm$1.2 & 48.3$\pm$0.8 \\
DEM~\cite{LiDEM} 
& 23.6$\pm$0.7 & 49.5$\pm$0.4 \\
GCN~\cite{PallabiGCN} 
& 22.3$\pm$0.6 & 49.7$\pm$0.6 \\
ER-ZSAR~\cite{ShizheERZSAR} 
& 42.1$\pm$1.4 & 73.1$\pm$0.3 \\
\midrule
X-CLIP-B/16~\cite{niXCLIP} 
& 65.2$\pm$0.4 & 86.1$\pm$0.8 \\
ASU-B/16~(ours)
& \textbf{67.6$\pm$0.2} & \textbf{87.2$\pm$0.3} \\
\bottomrule
\end{tabular}
\vspace{-1em}
\end{table}

\subsection{Ablation Study}
\label{sec:ablation_study}
We use the ASU-B/$16_{8f}$ with single-view inference for all the ablation experiments.  By default, the fully-supervised experiments are evaluated on Kinetics-400, the few-shot experiments are conducted on the first split of HMDB-51, and the zero-shot evaluation is on the first split of the validation set of UCF-101. 
\vspace{-0.4em}
\paragraph{Effects of Proposed Components}
We first investigate the effectiveness of the proposed components. 
The experiments are conducted under fully-supervised learning of Kinetics-400. 
(1) We treat the ActionCLIP-B/16-Transf, which introduces a 6-layer self-attention transformer followed by a mean pooling layer stacking on top of the CLIP image encoder as the baseline.
Results are presented in Tab.~\ref{tab:components}, and the first row denotes the baseline. 
(2) Compared with the baseline, replacing the transformer of baseline with our video decoder and introducing semantic units to generate queries for cross attention bring a \emph{0.9\%} gain in term of the top-1 accuracy.
The results demonstrate the effectiveness of enhancing visual representations with semantic embedding. 
(3) To validate the effectiveness of region-level representations, we disable the semantic-guided enhancement branch by feeding randomly initialized query embedding into the video decoder and retraining the model. The region-level representation can improve the top-1 accuracy of baseline by \emph{0.7\%}. 
(4) Combining the region-level representation and semantic unit further improve the value by \emph{0.6\%}.
The results indicate using proper region-level representation as enhancement can help alignments about instances and texts.
(5) Besides, equipping temporal relations in video decoder can further boost the top-1 accuracy by \emph{0.2\%}.
Overall, our proposed components can raise the top-1 accuracy of the baseline from \emph{81.1\%} to \emph{82.6\%}.
\vspace{-1.em}

\begin{table}
\caption{Ablation study on the effect of proposed components. \textit{Semantics}: semantic units. \textit{Region}: region-level representation. \textit{Temporal}: 1D-convolution layer in video decoder. }
\label{tab:components}
\centering
\resizebox{0.96\linewidth}{!}{
\begin{tabular}{lcccc}
\toprule
Semantic. & Region. & Temporal. & Top-1 Acc.(\%) & GFLOPs  \\
\midrule
-    & -    & -                               
& 81.1 & 141\\
\checkmark & - & -                           
& 82.0 & 138\\
- & \checkmark & -                  
& 81.8 & 145\\
\checkmark & \checkmark & -
& 82.4 & 146\\
\checkmark & \checkmark & \checkmark
& \textbf{82.6} & 146 \\
\bottomrule
\end{tabular}}
\vspace{-1.em}
\end{table}

\paragraph{Investigation of Semantic Units}
To validate the effect of semantic units, we first construct a baseline model, which replaces the queries to the video decoder with visual features from the region-aware image encoder.
The results are reported in the first row in Tab.~\ref{tab:semantics}. 
Then we investigate the effectiveness of each sub-collection in semantic units as defined in Eq.~\ref{action:labels}. 
From Tab.~\ref{tab:semantics}, we can observe the models equipped with semantic units improves the baseline, especially for the 2-shot (+\emph{5.1\%}) and 0-shot (+\emph{10.1\%}) experiments. 
The results indicate that visual representations without semantic guidance are not robust when facing a severe lack of data. 
We also conduct experiments to evaluate the four types of semantic 
units. In particular, the set of ``objects" is the most helpful to the results, indicating the objects are discriminative factors in the action recognition. 
The results in rows 2,3,4,5 show that each part of semantic units can bring performance gain, and their combination (in row 6) can achieve the best results. 
The phenomenon indicates that different types of semantic units complement each other, and languages benefit action recognition performance, especially in the transferability of the model.

\begin{table}
\caption{Ablation study on different types of semantic units.}
\label{tab:semantics}
\centering
\resizebox{0.75\linewidth}{!}{
\begin{tabular}{lccc}
\toprule
Semantic Units                         
& Fully. & 2-shot & 0-shot \\
\midrule
$\varnothing$                               
& 82.0 & 52.6 & 64.9 \\
$\mathcal{S}^{body}$                           
& 82.2 & 56.5 & 72.6 \\
$\mathcal{S}^{object}$                  
& 82.4 & 57.1 & 73.8 \\
$\mathcal{S}^{scene}$           
& 82.2 & 56.6 & 73.6 \\
$\mathcal{S}^{motion}$
& 82.1 & 56.3 & 72.2 \\
$\mathcal{S}$
& \textbf{82.6} & \textbf{57.7} & \textbf{75.0} \\
\bottomrule
\end{tabular}
}
\end{table}
\vspace{-1em}

\paragraph{Effects of Different Region Configurations}We conduct experiments with different spatial configurations to exploit spatial information for better action recognition. In particular, we split the whole image into regions according to two aspects: the number of split regions and the approach of spatial slicing.
Results are presented in Tab.~\ref{tab:regions}. There are two observations.
(1) Increasing the number of partitions will not continuously improve performance. When the size of each region is too large, the visual representations cannot align with the semantic units of small targets. 
On the contrary,  when the size of each region is too small, the receptive field of each region cannot cover the visual appearance of most semantic units. 
(2)  Vertical divisions cannot bring gain to the performance. We propose an assumption for the phenomenon. In an action video, the variations of the body parts usually happen in horizontal regions so that the horizontal regions can keep the majority of part structures across the frames, which can help better describe an action. In comparison, the vertical divisions will break such structures, leading to a performance decrease.

\begin{table}\small
\caption{Ablation study on the effect of different region configurations. \#Region is the number of regions. The configuration in Split denotes the horizontal number $\times$ the vertical number.}
\label{tab:regions}
\centering
\resizebox{0.97\linewidth}{!}{
\begin{tabular}{p{0.25\linewidth}<{\centering}p{0.03\linewidth}<{\centering}p{0.03\linewidth}<{\centering}p{0.03\linewidth}<{\centering}p{0.03\linewidth}<{\centering}p{0.03\linewidth}<{\centering}p{0.03\linewidth}<{\centering}}
\toprule
\# Region & 0 & 2 & \multicolumn{2}{c}{4}& \multicolumn{2}{c}{8} \\
Split & - & 2$\times$1 & 4$\times$1  & \multicolumn{1}{c}{2$\times$2} & 8$\times$1  & \multicolumn{1}{c}{4$\times$2} \\
\midrule
Top1-Acc(\%). & \multicolumn{1}{l}{82.0} & \multicolumn{1}{l}{82.3} & \multicolumn{1}{l}{\textbf{82.6}} & 82.2  & \multicolumn{1}{l}{82.4} & 81.8 \\
\midrule
GFLOPs & \multicolumn{1}{l}{138} & \multicolumn{1}{l}{140} & \multicolumn{1}{l}{146} & 146  & \multicolumn{1}{l}{162} & 162 \\
\bottomrule
\end{tabular}
}
\vspace{-1.5em}
\end{table}


\section{Discussion}
\noindent \textbf{Conclusion} In this paper, we propose a new video representation learning framework for action recognition based on pre-trained visual-language models. 
To leverage fine-grained visual-language correlation, we introduce ``semantic units" that provide factorized and reusable textual knowledge hiding behind actions.
A multi-region attention module that perceives region-aware information is proposed to better capture fine-grained alignments. Under the guidance of selected semantic units, a cross-modal decoder is designed for decoding spatiotemporal video representations.
Our approach produces discriminative video representations, outperforming the state-of-the-art approaches in fully supervised comparisons. The semantic units' re-usability further alleviates the learning difficulty of adapting our model to new scenarios, achieving superior results in the zero-shot/few-shot settings.

\noindent \textbf{Limitations and Future Work}
There are two potential improvements for ASU to explore in the future. Firstly, the generation of semantic units still requires a few manual operations. With the increasing maturity of large-scale language models such as the GPT series~\cite{BrownGPT2020,OuyangInstructGPT2022}, the current procedure can be fully automated with appropriate prompts and CoT~\cite{WeiCoT2022} designs. Secondly, ASU and other recent VLM-based approaches consume significant computational resources on pre-trained backbones during the inference phase. From the perspective of practical application, incorporating more computational-efficient techniques is a valuable research field.


{\small
\bibliographystyle{ieee_fullname}
\bibliography{ASU_arxiv}
}
\appendix
\section{Architecture Details}\label{Architecture Details}
\noindent \textbf{Semantic Units.} We introduce 366 semantic units for the Kinetics-400 dataset according to its label names. Specifically, these units consist of 23 body parts, 232 objects, 51 scenes, and 60 motion units. Similarly, we collect 556, 100, and 60 semantic units for Kinetics-600, UCF-101, and HMDB-51, respectively. 
In few/zero-shot experiments, these semantic units could be leveraged together as a common bank. 
Fig.~\ref{fig:semantic_units_generation} demonstrates a snapshot of our standard semantic unit generation procedures. It is noteworthy that during procedures, identical semantic units may repeatedly emerge due to their reusability across different labels and datasets. We filter out the duplicated ones in case of disturbance to the semantic query generator module.  

\setcounter{figure}{0}
\renewcommand{\thefigure}{A\arabic{figure}} 
\begin{figure*}[t]
\centering
\includegraphics[width=0.95\linewidth]{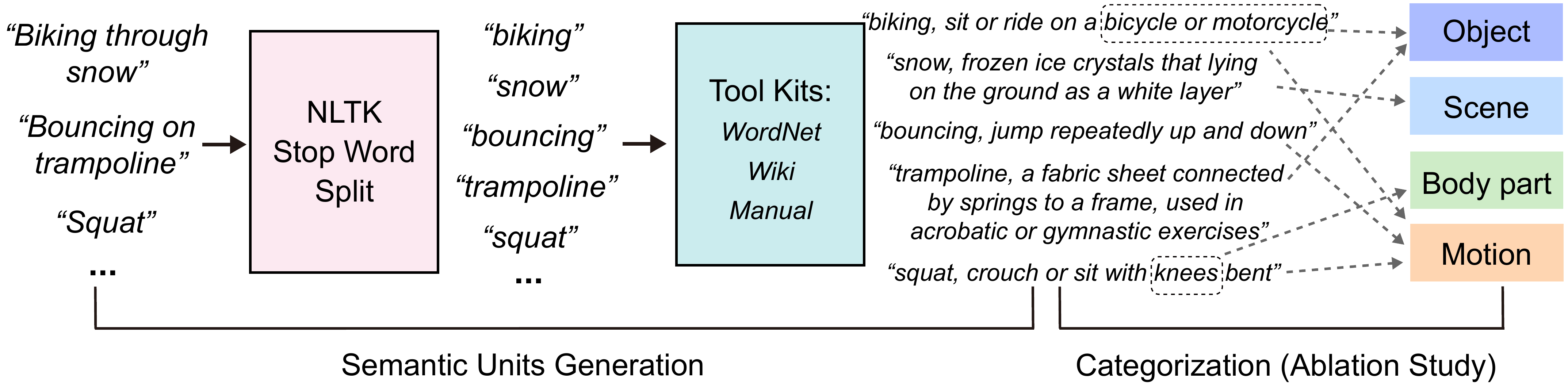}
\caption{Snapshot of semantic units generation
}
\label{fig:semantic_units_generation}
\end{figure*}

\noindent \textbf{Proposed Network.} ASU-B/16 employs CLIP-B/16~\cite{radfordLearningTransferableVisual2021} ($L$=12, $N$=12, $d$=768, $p$=16) as the text encoder and ViT backbone of the region-aware image encoder, while ASU-L/14 adopts CLIP-L/14 ($L$=24, $N$=16, $d$=1024, $p$=14), where $L$ denotes the layers, $N$ denotes the number of attention heads, $d$ refers to the embedding dimension and $p$ indicates the patch size. It is noteworthy that ASU-L/14@336 takes frames with resolution 336$\times$336 while others adopt the resolution 224$\times$224.  The number of cross-attention (CA) modules in the region-aware image encoder and the video decoder is 2 and 4, respectively. The semantic query generator is parameter-free, and the softmax temperature is set to 0.01. For 1D-convolution layers in the video decoder, we set the kernel size to 3, stride to 1, and the number of groups to the number of input channels.

\section{Hyperparameters and Training Details} \label{Hyperparameter Details}
In fully-supervised training, we set the batch size to 256 and adopt the AdamW optimizer with $\beta_1=0.9$ and $\beta_2=0.98$. The learning rate is 8$\times10^{-6}$ for the ViT module in the region-aware image encoder and 8$\times10^{-5}$ for the remaining learnable parts. In few-shot experiments, the learning rate is scaled up by ten times, and the batch size is set to 64. We train our model for 30 epochs with a weight decay of 0.001. The learning rate is warmed up for five epochs and decayed with a cosine schedule. Training is conducted on a server with eight 32GB-V100-GPUs. For data augmentation, we utilize the technique including \emph{RandomFlip, MultiScaleCrop, Mixup, and Label smoothing}, following the manner of X-CLIP~\cite{niXCLIP}.

\setcounter{table}{0}
\renewcommand{\thetable}{B\arabic{table}} 
\begin{table}[t]
\centering
\caption{Analysis on text prompting methods}
\label{Tab:text_prompt}
\setlength{\tabcolsep}{8mm}
\resizebox{0.95\linewidth}{!}{
\begin{tabular}{lc}
\toprule
Prompt Method & Top-1 Acc.(\%)  \\ \hline
w/o Prompt   & 82.4       \\  
Fixed Template & \textbf{82.6}   \\
Templates Ensemble  & \textbf{82.6}   \\ 
CoOp & 82.3\\
CoCoOp    & 82.2      \\ 
Video-specific Prompt & \textbf{82.6} \\
Learnable Vectors & 82.5\\
\bottomrule
\end{tabular}
}
\vspace{-1em}
\end{table}

\section{Additional Experiments Analysis} \label{sec:additionalExp}
\subsection{Analysis on Text Prompting Methods}
We investigate the effect of different prompting approaches on the text encoder branch. These approaches include traditional template prompting techniques and recently proposed learnable prompting methods: (1) Feeding original class name into the text encoder. (2) Fixed Template: applying a fixed template, "a video of a person \{label\}." to enhance the content cues in videos. (3) Templates Ensemble: applying fixed hand-craft template set from ~\cite{wangActionclipNewParadigm2021} on class labels. Templates are randomly chosen in each training iteration, and the results in the inference stage are the average of all the templates. (4) CoOp: adopting the technique of ~\cite{ZhouCoOp22}, which adding learnable vector space with length=16 to tokenized text labels. (5) CoCoOp: further adding a meta-net~\cite{ZhouCoCoOp22} incorporating visual information into the vector space. (6) Visual-specific prompt: employing a 2-layer cross attention module ~\cite{niXCLIP} on top of the text encoder. It uses the video representation as queries and text representation as keys\&values. (7) Learnable vectors: introducing a learnable vector W that is initialized by the CLIP's text features of labels. We replace the text features of labels with the vectors in the training pipeline. The learning rate is set to 1$\times10^{-4}$, and we regularize the vectors by adding $L_2$ loss $L_{reg}=||$W$ - $C$||_2$, where C denotes the text features of action labels. We leverage these approaches on ASU-B/16 separately and train the models on Kinetics-400. The top-1 accuracies are presented in Tab.~\ref{Tab:text_prompt}. We find that different templates do not significantly impact our model, and using a fixed template is an efficient method compared to methods that introduce extra learnable modules.

\subsection{Analysis on Loss Functions}
Our proposed method supervises video representations with text features. Despite the inherent open-vocabulary advantage, we investigate the superiority of leveraging such a cross-modal loss in fully-supervised experiments. The experiments are conducted in the following settings: (1) We impose supervision under a uni-modal loss (one-hot label) function by introducing a classification head on top of the video encoder. The head only takes the video representation $\mathbf{z} \in \mathbb{R}^{d}$ and outputs the single-modal logits $\mathbf{g} \in \mathbb{R}^{N}$ for N action labels, which is used to calculate the cross-entropy loss with the ground truth. (2) We adopt our proposed cross-modal loss function. Specifically, the loss function is implemented by the cross entropy of cross-modal logits $\mathbf{C}\mathbf{z}$, where $\mathbf{z}$ denotes the video representation and $\mathbf{C} \in \mathbb{N}^{d}$ represents the text features of N action labels. (3) We further utilize both losses as an ensemble one. We 
employ ASU-B/16 and train the models on Kinetics-400. The top-1 accuracies are reported in Tab.~\ref{Tab:loss_and_modal}. The result shows that adopting cross-modal supervision in our proposed network achieves better performance, which validates the strength of the vison-language framework. Meanwhile, utilizing ensemble losses further gains +0.1$\%$. We attribute such a complementary effect to the regularization functionality of the uni-modal classification branch. For instance, the text features of \texttt{drinking beer} and \texttt{drinking shot} are relatively close in the feature space. Leveraging them as guidance may make learning distinctions among the corresponding labels difficult. Introducing uni-modal classification, which treats labels as uniform one-hot vectors, can alleviate the problem.         

\setcounter{table}{0}
\renewcommand{\thetable}{C\arabic{table}} 
\begin{table}[t]
\centering
\caption{Analysis on loss functions}
\label{Tab:loss_and_modal}
\setlength{\tabcolsep}{8mm}
\begin{tabular}{lr}
\toprule
Loss Type & Top-1 Acc.(\%)  \\ \hline
Uni modality   & 81.9       \\  
Cross modality & 82.6   \\
Ensemble  & \textbf{82.7}   \\ 
\bottomrule
\end{tabular}
\end{table}

\section{Qualitative Results} 
\label{sec:QualitativeResults}
\subsection{Visualization of Semantic Attention Weights}
We provide qualitative results about the semantic query generator's attention weights $a_{i,j}$ for some sampled frames. The attention weights reflect affinities between the visual representations and the semantic units. The semantic units with higher attention scores play more significant roles in video representation learning. In Fig.~\ref{fig:attention_weights}, we show eight sampled frames per video along with a stacked bar plot representing the associated affinity scores to semantic units. For each frame, there are five stacked bars of the attention weights belonging to the frame-level features and the four region-level ones separately. For better visualization, we show the scores of top-3 semantic units (without discriminative description) and scale their sum to 1. For example, the instance in the top row shows a video segment of action smoking hookah. We can see that frame and region features correctly pay more attention to the semantic unit \texttt{Hookah, a smoking tool} in the middle frames, where the object starts to appear. The phenomenon reveals that our proposed semantic query generator can extract effective semantic cues to supervise video representation learning. Besides, attention weights of different regions are distinguished due to the motion of concerned units. We propose that such variance can help the model explore the spatiotemporal information of videos.

\setcounter{figure}{0}
\renewcommand{\thefigure}{C\arabic{figure}} 
\begin{figure*}[t]
    \centering
    \includegraphics[width=0.92\textwidth]{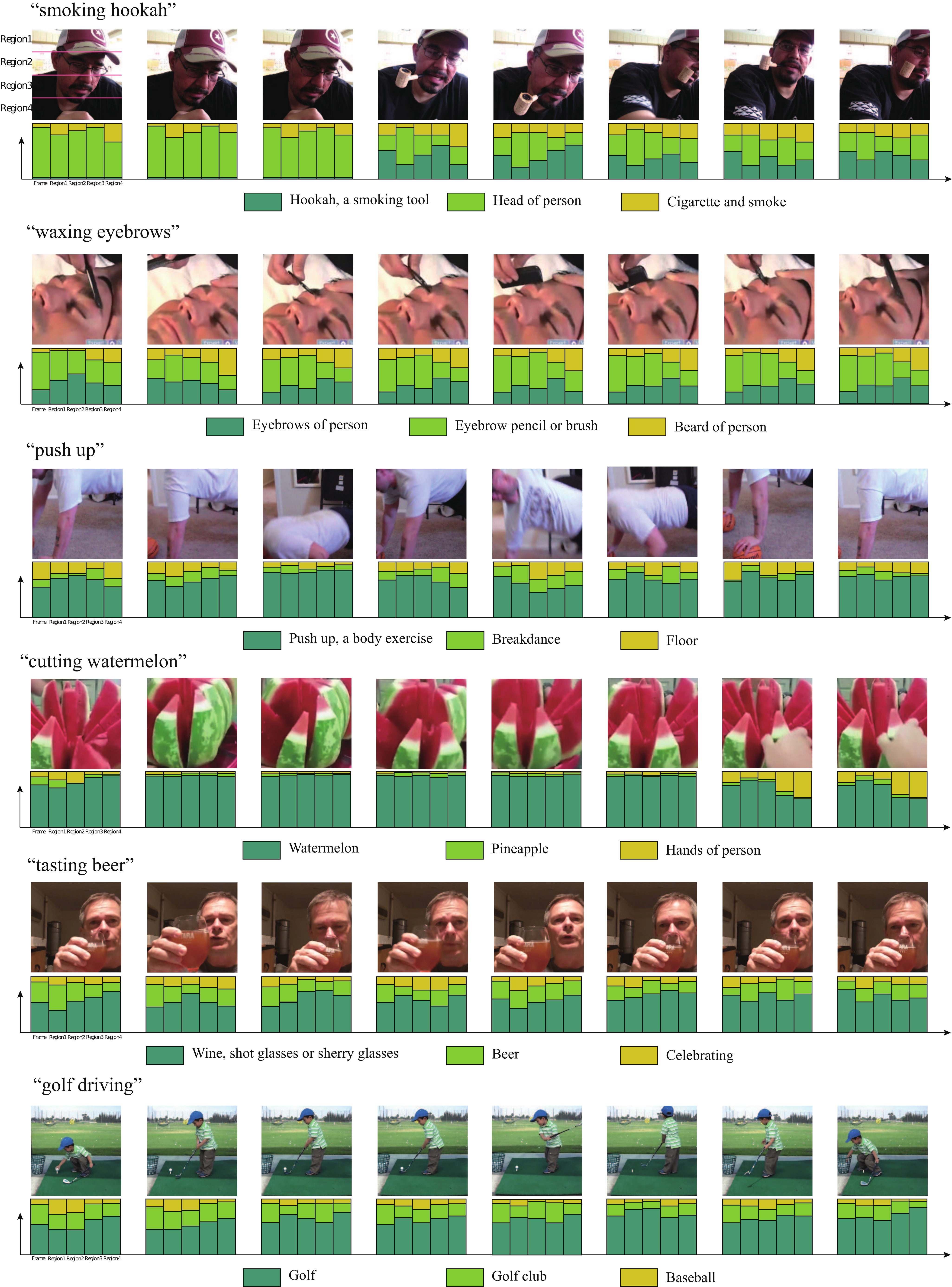} 
    \caption{Visualization of semantic attention}
    \label{fig:attention_weights}  \vspace{-1em}
\end{figure*}

\begin{figure*}[t]
    \centering
    \includegraphics[width=0.78\textwidth]{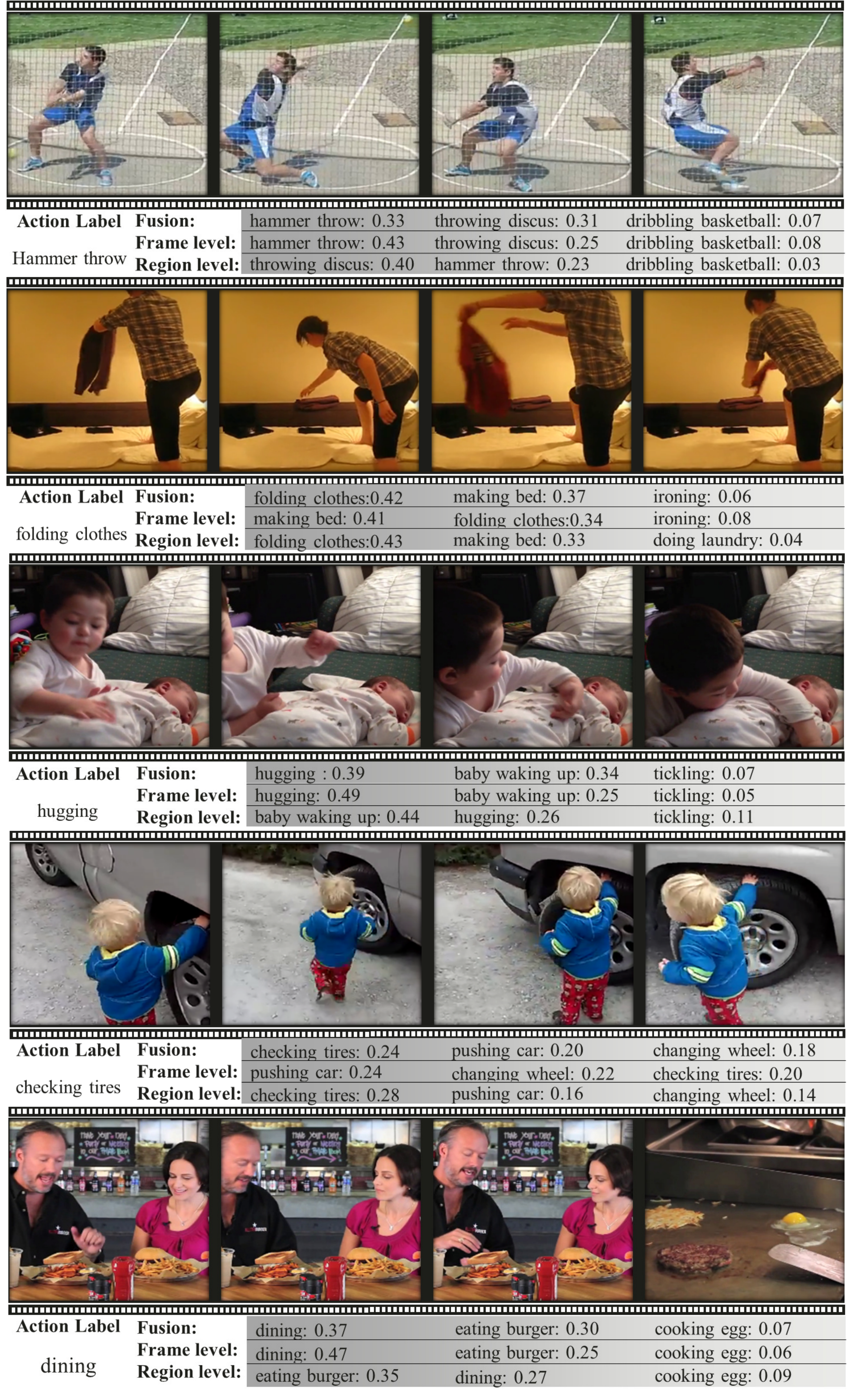}  \vspace{-1em}
    \caption{Visualization of Inference}
    \label{fig:vis_frame_region}  \vspace{-1em}
\end{figure*}

\subsection{Visualization of Inference Results}
In this section, we demonstrate the qualitative results of our proposed network's inference. To validate the effectiveness of feeding frame-level and region-level visual representations described in Sec~\ref{sec:ablation_study}, we train ASU-B/16 with three different compositions of visual features fed into the video decoder: frame-level features only, region-level features only, and fusion features (our proposed methods) that concatenate both. As shown in Fig.~\ref{fig:vis_frame_region}, we compare the inference results of the same video under three settings, and top-3 predictions are presented. We observe a complementary effect of utilizing multi-scale information, namely frame-level and region-level features. For instance, the first example is wrongly predicted as throwing discus under region-level features settings, while the model that absorbs frame-level features can correctly categorize it as an act of throwing a hammer. On the contrary, the second example, where a person is folding clothes, can be recognized well with the help of region-level features.

\subsection{Failure cases analysis}
Our method `selects' relevant semantic units and establishes semantic guidance for frames through a softmax-weight-sum manner semantic attention module (Fig.~\ref{fig:query_gen_vid_dec}b), which means each input frame corresponds to the whole set of semantic units with ``attention scores." We observe a few `failure' cases that undesirable semantic units get higher scores. When inferring a video of action ``salsa dancing," some sampled frames give higher scores on irrelevant units like ``tango" and ``swing" other than ``salsa."  We investigate the corresponding video samples of these units and find that they are hard to distinguish merely from a static frame.

\end{document}